# An Efficient Preprocessing Methodology for Discovering Patterns and Clustering of Web Users using a Dynamic ART1 Neural Network

Ramya C*, Kavitha G**
*Department of Studies in Computer Science and Engineering
U.B.D.T. College of Engineering, Davangere University, Davangere-04
**Lecturer, U.B.D.T.C.E, Davangere University, Davangere-04
cramyac@gmail.com

**Abstract :** In this paper, a complete preprocessing methodology for discovering patterns in web usage mining process to improve the quality of data by reducing the quantity of data has been proposed. A dynamic ART1 neural network clustering algorithm to group users according to their Web access patterns with its neat architecture is also proposed. Several experiments are conducted and the results show the proposed methodology reduces the size of Web log files down to 73-82% of the initial size and the proposed ART1 algorithm is dynamic and learns relatively stable quality clusters.

## 1 INTRODUCTION

Web log data is usually diverse and voluminous. This data must be assembled into a consistent, integrated and comprehensive view, in order to be used for pattern discovery. Without properly cleaning, transforming and structuring the data prior to the analysis one cannot expect to find the meaningful patterns. Rushing to analyze usage data without a proper preprocessing method will lead to poor results or even to failure. So we go for preprocessing methodology. The results show that the proposed methodology reduces the size of Web access log files down to 73-82% of the initial size and offers richer logs that are structured for further stages of Web Usage Mining (WUM).

We also present an ART1 based clustering algorithm to group users according to their Web access patterns. In our ART1 based clustering approach, each cluster of users is represented by a prototype vector that is a generalized representation of URLs frequently accessed by all the members of that cluster. One can control the degree of similarity between the members of each cluster by changing the value of the vigilance parameter. In our work, we analyze the clusters formed by using the ART1 technique by varying the vigilance parameter ρ between the values 0.3 and 0.5.

## 2 PREPROCESSING METHODOLOGY

The main objectives of preprocessing are to reduce the quantity of data being analyzed while, at the same time, to enhance its quality. Preprocessing comprises of the following steps – Merging of Log files from Different Web Servers, Data cleaning, Identification of Users, Sessions, and Visits, Data formatting and Summarization as shown in Fig. 1.

### 2.1 Merging
At the beginning of the data preprocessing, the requests from all log files, put together into a joint log file with the Web server name to distinguish between requests made to different Web servers and taking into account the synchronization of Web server clocks, including time zone differences.

### 2.2 Data Cleaning
The second step of data preprocessing consists of removing useless requests from the log files. Since all the log entries are not valid, we need to eliminate the irrelevant entries. Usually, this process removes requests concerning non-analyzed resources such as images, multimedia files, and page style files.

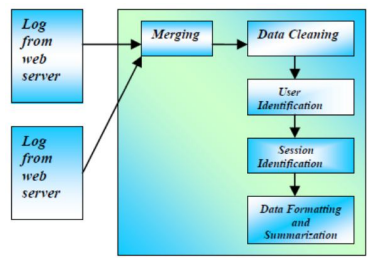

**Fig. 1.** Stages of Preprocessing

### 2.3 User Identification
In most cases, the log file provides only the computer address (name or IP) and the user agent (for the ECLF log files). For Web sites requiring user registration, the log file also contains the user login (as the third record in a log entry) that can be used for the user identification.

### 2.4 Session Identification
A user session is a directed list of page accesses performed by an individual user during a visit in a Web site A user may have a single (or multiple) session(s) during a period of time. The session identification problem is formulated as "Given the Web log file, capture the Web users' navigation trends, typically expressed in the form of Web users' sessions".

### 2.5 Data Formatting & Summarization

This is the last step of data preprocessing. Here, the structured file containing sessions and visits are transformed to a relational database model.

## 3 CLUSTERING METHODOLOGY

A clustering algorithm takes as input a set of input vectors and gives as output a set of clusters thus mapping of each input vector to a cluster. A novel based approach for dynamically grouping Web users based on their Web access patterns using ART1 NN clustering algorithm is presented in this paper. The proposed ART1 NN clustering methodology with a neat architecture is discussed.

### 3.1 The Clustering Model

The proposed clustering model involves two stages – Feature Extraction stage and the Clustering Stage. First, the features from the preprocessed log data are extracted and a binary pattern vector $P$ is generated. Then, ART1 NN clustering algorithm for creating the clusters in the form of prototype vectors is used. The feature extractor forms an input binary pattern vector $P$ that is derived from the base vector $D$. The procedure is given in Fig. 2. It generates the pattern vector which is the input vector for ART1 NN based clustering algorithm.

```
Procedure Gen_Pattern ( )
Begin
  for each pattern vector P_H, where H = 1 to n
  for each element p_i in pattern vector P_H, i=1 to m
    if URL_i requested by the host more than twice
      then p_i = 1;
      else p_i = 0;
        End
```

**Fig. 2.** Procedure for Generating Pattern Vector

The architecture of ART1 NN based clustering is given in Fig. 3. Each input vector activates a winner node in the layer F2 that has highest value among the product of input vector and the bottom-up weight vector. The F2 layer then reads out the top-down expectation of the winning node to F1, where the expectation is normalized over the input pattern vector and compared with the vigilance parameter $\rho$. If the winner and input vector match within the tolerance allowed by the $\rho$, the ART1 algorithm sets the control gain G2 to 0 and updates the top-down weights corresponding to the winner. If a mismatch occurs, the gain controls G1 & G2 are set to 1 to disable the current node and process the input on another uncommitted node. Once the network is stabilized, the top-down weights corresponding to each node in F2 layer represent the prototype vector for that node. Summary of the steps involved in ART1 clustering algorithm is shown in Table 1.

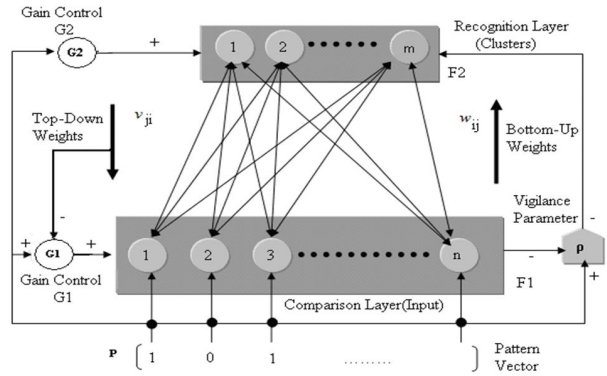

**Fig. 3.** Architecture of ART1 NN based Clustering

**Table 1.** Algorithm of ART1 NN Clustering

1. Initialize the vigilance parameter $\rho$, $0 \leq \rho \leq 1$, $w = 2/(1+n)$, $v=1$ where $w$ is $m \times n$ matrix (bottom-up weights) and $v$ is the $n \times m$ matrix (top-down weights), for $n$-tuple input vector and $m$ clusters.
2. Binary unipolar input vector $p$ is presented at input nodes. $p_i=\{0,1\}$ for i=1,2,..,n
3. Compute matching scores

$$y_k^0 = \sum_{i=1}^{n} w_{ik} p_i \text{ for k=1,2,...m}$$

Select the best matching existing cluster $j$ with $y_j^0 = Max(y_k^0), k = 1,2,..,m$

4. Perform similarity test for the winning neuron

$$\frac{\sum_{i=1}^{n} v_{ij} p_i}{\|p\|_1} > \rho$$

Where $\rho$, the vigilance parameter and the norm $\|p\|_1$ is the $L_1$ norm defined as,

$\|p\|_1 = \sum_{i=1}^{n} |p_i|$, if the test is passed, the algorithm goes to step 5. If the test fails, then the algorithm goes to step 6, only if top layer has more than a single active node left otherwise, the algorithm goes to step 5.

5. Update the weight matrices for index $j$ passing the test (1). The updates are only for entries $(i,j)$ where i=1,2,..,m and are computed as follows:

$$w_{ij}(t+1) = \frac{v_{ij}(t) p_i}{0.5 + \sum_{j=1}^{n} v_{ij}(t) p_i} \quad \text{and} \quad v_{ij}(t+1) = p_i \, v_{ij}(t)$$

This updates the weights of $j^{th}$ cluster (newly created or the existing one). Algorithm returns to step 2.

6. The node $j$ is deactivated by setting $y_j$ to 0. Thus this node does not participate in the current cluster search. The algorithm goes back to step 3 and it will attempt to establish a new cluster different than $j$ for the pattern under test.

## 4  EXPERIMENTAL RESULTS

We have conducted several experiments on log files collected from NASA Web site during July 1995. Through these experiments, we show that our preprocessing methodology reduces significantly the size of the initial log files by eliminating unnecessary requests and increases their quality through better structuring. It is observed from the Table 2 that, the size of the log file is reduced to 73-82% of the initial size. The Fig. 4 shows the GUI of our toolbox with preprocessor tab and ART Clustering tab. The Preprocessor Tab allows the user to perform preprocessing operation. The ART Clustering tab allows the user to perform clustering operation. The pattern vector (corresponding to the preprocessed log file) and vigilance parameter values have to be specified as inputs. The prototype vectors of the clusters are displayed at the result panel. The results show that the proposed ART1 algorithm learns relatively stable quality clusters.

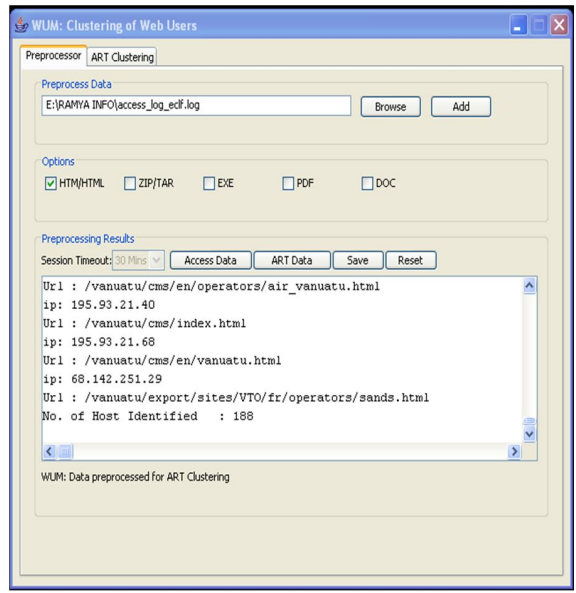

**Fig. 4.** WUM toolbox: Results after preprocessing (NASA Log file,1995)

## 5  CONCLUSIONS

In this paper, we have presented an effective methodology for preprocessing required for WUM process. The experimental results illustrate the importance of the data preprocessing step and the effectiveness of our methodology. Next, we present ART1 clustering algorithm to group hosts according to their Web request patterns. The experimental results prove that the proposed ART1 algorithm learns relatively stable quality clusters.

**Table 2**. Results after Preprocessing

| Website | Duration | Original Size | Size after Preprocessing | % Reduction in Size | No. of Sessions | No. of Users |
|---|---|---|---|---|---|---|
| NASA | 1-10th Aug 95 | 75361 bytes (7.6MB) | 20362 bytes | 72.98% | 6821 | 5421 |
| NASA | 20-24th July 95 | 205532 bytes (20.6MB) | 57092 bytes | 72.22% | 16810 | 12525 |
| Academic Site | 12-28th May 2001 | 28972 bytes (2.9MB) | 5043 bytes | 82.5% | 1645 | 936 |